# Multi-modal Video Chapter Generation

Xiao Cao  
Shandong University

Zitan Chen[1]  
Shandong University

Canyu Le[2]  
ByteDance

Lei Meng[3]  
Shandong University

**Abstract**

Chapter generation becomes practical technique for online videos nowadays. The chapter breakpoints enable users to quickly find the parts they want and get the summative annotations. However, there is no public method and dataset for this task. To facilitate the research along this direction, we introduce a new dataset called Chapter-Gen, which consists of approximately $10k$ user-generated videos with annotated chapter information. Our data collection procedure is fast, scalable and does not require any additional manual annotation. On top of this dataset, we design a effective baseline specificlly for video chapters generation task. which captures two aspects of a video,including visual dynamics and narration text. It disentangles local and global video features f or localization and title generation respectively. To parse the long video efficiently,a skip sliding window mechanism is designed to localize potential chapters. And a cross attention multi-modal fusion module is developed to aggregate local features for title generation. Our experiments demonstrate that the proposed framework achieves superior results over existing methods which illustrate that the method design for similar task cannot be transferred directly even after fine-tuning. Code and dataset are available at https://github.com/czt117/MVCG.

## 1 Introduction

Video chapter generation aims to understand the whole of video and conclude its content to a couple of short chapter titles, just like chapters in a book. This task is an important step towards building high-level semantic understanding of videos. It also offers many valuable industrial applications such as video preview, quick searching, content-driven video retrieval and video editing.

Over the past decade, extensive studies have been devoted to video understanding, e.g. action recognition and localization [4, 16, 17, 35], event detection [30] and video summarization [7, 29]. However, methods developed for these tasks are not applicable for video chapter generation. Specifically, the videos in these tasks consist of simple scenes and its durations are usually short. Visual features are sufficient for achieving good results. In contrast, the videos for chapter generation are longer and contain many complex temporal structures. Other high-level video understanding tasks like video captioning [24, 34] and dense video captioning [11, 14] are also different from our task. From the output aspect, those video captioning tasks try to generate narrative descriptions for some short video events. For example, video captioning may generate "A woman walks to the piano and briefly talks to the the elderly man" for a 15-seconds video event. While our proposed task generate conclusive chapter title or theme like "Memory Promoting Exercise" for 2 minutes video snippet. Some examples of our task are illustrated in Figure 1. From the model aspect, their methods model the "sematic actions" (i.e. characters' actions, scene changes) in scenes instead of the high-level summarization for chapters.

---

1. Contribute equally
2. Industrial corresponding author
3. Academic corresponding author

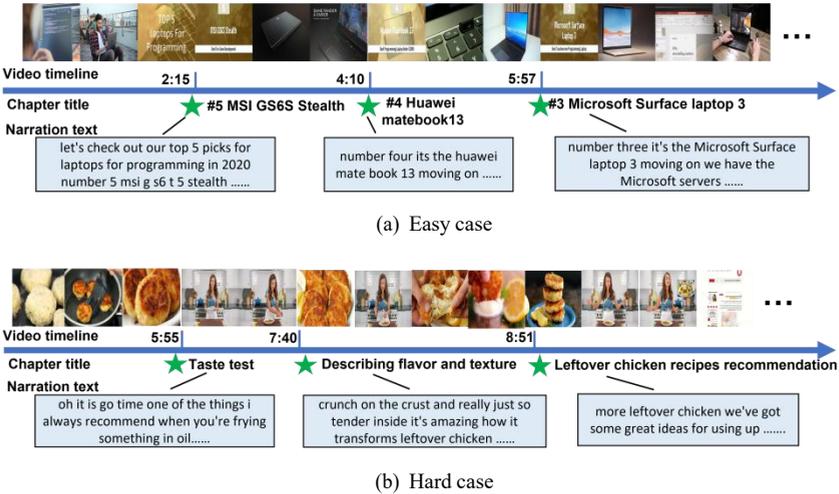

Figure 1: We show the easy and hard cases in our dataset. (a) shows an easy case in which each chapter has clear temporal structures. (b) shows a hard case. There are no clear visual or narration features to determine where the chapter beginning is without watching most parts of the video.

Recently, several movie datasets [2, 10] and movies scene methods [5, 22] are developed, which consist of many shots, scenes and stories. Movies could be made based on narrative scripts, but they lack conclusive chapter labels for video chapter generation task. Moreover, the movie scene analysis methods [5, 22] usually develop their approaches on a couple of keyframe sequences while those keyframe sequences are sparse and a lot of original information may lose. Such datasets and approaches could not be suitable for video chapter generation.

To facilitate the research in video chapter generation, a new large and diverse dataset is needed. More importantly, it should allow high-level semantics and clear temporal structures to be extracted and analyzed. Instead of manually annotation, in this work, we explore a different source of supervision to obtain the videos with labeled chapter information. We observe that plenty of user videos are available on YouTube. And some are with uploaded video chapter information which contain chapter beginning timestamps and chapter titles. These information are manually uploaded by video authors. To leverage this rich source of data, we construct a new dataset containing 9631 videos that depict various activities and topics including entertainment, education, business, etc (See Table 2).

On top of this new dataset, we develop a new framework to generate video chapters. Rather than processing extracted keyframes or trimmed fragments in [22], we consider dense video frames, which contain more complete video context for chapter generation. Specifically, our framework contains two stages: (1) **Chapter localization.** (2) **Chapter title generation.**

Overall, the two-stage framework addresses the chapter localization and chapter title generation, respectively, and disentangles local and global video features for two sub-tasks. The experiments demonstrate that our two-stage frame works well and surpasses other existing competitors.

In summary, our contributions lie in three aspects:

- We introduce a novel high-level video understanding task: video chapter generation.
- We construct a new dataset called ChapterGen, which contain annotated associations between chapter beginning timestamps and chapter titles. We leverage readily available resources. Therefore, the data collection is fast, scalable and does not require expensive manual annotation. This dataset can be used as a standard benchmark for video chapter study, which we believe is a promising direction towards high-level video understanding.
- We develop a two-stage framework to address the video chapter generation problem.

## 2 Related Work

### 2.1 Video understanding dataset

In recent years, with the increasing interest in video understanding, many video understanding tasks and datasets have been proposed. The low-level video understanding datasets like Youtube-8M [1] for video classification, Activitynet [3], Kinetics-400 [13] for action recognition. These datasets have plentiful short or trimmed videos consisting of several action sequences. For high-level video understanding tasks, TVSum [29] contains 50 video sequences for video summarization, MPII-MD [24] has 94 videos for video captioning. ActivityNet Captions dataset[15] has 20k videos for dense video captioning (DVC). But the duration of a chapter in our dataset is about three times longer than the event duration in DVC dataset. And the length of videos in our dataset is around six times longer than theirs. The MovieNet dataset [9] provides 1100 movies, partially annotated with story scene boundaries, cinematic styles and movie synopses. More recently, some works also introduce large-scale datasets using readily available resources for video language representation learning [31, 37] and text-video retrieval [20]. These datasets contain many video clip-caption pairs for self-supervised learning without explicit annotations. The comparison between our dataset and existing high-level video understanding datasets is summarized in Table 1.

### 2.2 Video temporal localization

Video temporal localization is the problem of identifying some target locations in videos with its beginning and end times. This problem covers a wide range of tasks such as video action temporal localization, video shot detection, movie scene segmentation[5, 22] and dense video captioning[11, 12, 14] . Video action temporal localization [17, 18, 26] is to detect some specific actions start and end from the dense frame sequences. Video shot detection is to find shots based on some low-level features, including color histogram [25], camera motion parameters [23] and audio clues [27].

| Dataset | Video *num.* | Length (min.) | Task |
|---|---|---|---|
| ActivityCaptions [15] | 20k | 2.1 | Dense video captioning |
| TVSum [29] | 50 | 4.2 | Video summarization |
| MPII-MD [24] | 94 | 26.1 | Video captioning |
| MovieNet [10] | 1100 | 118.8 | Movie analysis |
| MovieScenes [22] | 150 | 118.8 | Movie scene detection |
| ChapterGen | 9631 | 13.6 | Video chapter generation |

Table 1: The comparison between ChapterGen and existing high-level video understanding datasets which contain explicit annotations.

## 2.3 Text summarization

Given a long text, text summarization is to understand the meaning of the provided long text and output concise and conclusive sentences. The text summarization approaches can be categorized as extractive summarization [21, 39] and abstrastive summarization [28, 38]. Extractive summarization which directly copies informative fragments from the input, while abstractive summarization attempt to generate new words or sentences.

# 3 ChapterGen Dataset

We construct a new dataset ChapterGen on 9631 videos which consist of various topics like entertainment, education, business, daily life, etc. The data collection does not rely on expensive annotations and one can easily expand ChapterGen to larger data scale.

## 3.1 Data Collection

We notice that plentiful videos with chapter information are available on YouTube. However, the videos with chapters or without chapters are mixed and there is no obvious flag to distinguish them. Inspired by a recent video dataset [20], we first obtain a large list of activities from WikiHow. Unlike [20] excluding abstract topics, we would like to cover various topics as many as possible. Hence, we introduce other topics like leaderboard, interviews, round-table discussion after including all categories in WikiHow.

After acquiring the list of topic activities, we use them to automatically search for related YouTube videos via official YouTube API. For each topic activity, we restrict to the top 200 search results and discard videos without chapter information by checking the timestamps in video description.

In addition, we downsample them by 1 frame/second but keep all narrations for a video contains a lot of duplicated or same frames. In this way, the model can quickly parse the whole video without loss of chapter generation accuracy. We also intentionally discard videos whose duration is more than 30 minutes, so that model can be efficiently trained on more different videos. Note that our approach is general and not limited by video duration. It is able to directly handle longer videos (e.g. a couple of hours) without any modification.

| Category | Video num. | Chapter dur.(s) | Chapter num. | Title len. | Category | Video num. | Chapter dur.(s) | Chapter num. | Title len. |
|---|---|---|---|---|---|---|---|---|---|
| Arts and Entertainment | 846 | 95.16 | 8.76 | 3.79 | Personal Care and Style | 493 | 87.91 | 9.01 | 3.88 |
| Cars and Vehicles | 491 | 95.92 | 8.50 | 3.71 | Pets and Animals | 310 | 94.97 | 8.51 | 3.95 |
| Health | 613 | 88.78 | 8.50 | 3.90 | Family Life | 442 | 102.79 | 8.53 | 4.18 |
| Finance and Business | 674 | 96.90 | 8.43 | 4.24 | Work World | 217 | 89.35 | 9.06 | 4.19 |
| Sports and Fitness | 604 | 91.27 | 8.36 | 3.98 | Holidays and Traditions | 175 | 101.21 | 8.63 | 4.08 |
| Hobbies and Crafts | 435 | 98.76 | 8.64 | 3.76 | Travel | 188 | 89.63 | 9.64 | 3.82 |
| Education and Communications | 579 | 95.72 | 8.57 | 3.92 | Youth | 402 | 96.50 | 9.08 | 4.59 |
| Food and Entertaining | 435 | 89.11 | 8.55 | 3.66 | Others | 1812 | 91.68 | 8.95 | 3.98 |
| Computers and Electronics | 570 | 82.49 | 9.38 | 3.76 | Home and Garden | 345 | 93.27 | 9.12 | 3.75 |
| Total | - | - | - | - | Total | 9631 | - | - | - |

Table 2: The statistics of ChapterGen dataset. Video categories, The number of video, average chapter duration, average chapter number per video and average title length per chapter are reported respectively. Others category includes leaderboard, interviews, round-table discussion which are not covered by WikiHow.

Since videos are automatically collected, there are some anomaly data. We further improve the quality and consistency of the dataset by applying the following criteria. (1) We discard videos whose duration is less than 100 seconds. Because the short video contains insufficient information to generate valid chapters. (2) We also rule out videos in which the chapter duration (i.e. the gap between two consecutive chapters) is less than 8 seconds. If a chapter is too short, it is either incorrectly written by an amateurish video author or lacks useful information for chapter generation.

## 3.2 Dataset Statistics

The video category distribution and chapter properties are reported in Table 2. In chapterGen, each video have 9 chapters in average. Each chapter duration is roughly 90 seconds with 4 words chapter title. Moreover, we observe that there is no uniform rule or guidance when video authors write their chapter information. Some videos have clear content and temporal structures. But some have relatively vague structures. To evaluate performance thoroughly, we manually inspect the collected videos and mark each video as an easy or hard case. This annotation is done by 12 annotators. The easy case represents the video has clear structures. It is easy to distinguish chapter boundaries and other video content. The hard case means that the video has vague structures and there are no obvious visual or narration features to identify chapter boundaries. Figure 1 gives some examples to demonstrate this concept. There are total 5861 easy cases and 2102 hard cases. The rest of data are kind of ambiguous: some chapters in a video can be seen as easy cases but others are hard cases.

## 4 Methodology

In this section, we present our framework to generate video chapters. Formally, in ChapterGen dataset, a video contains a series of image frames $X$ and narration text $S$. The chapter information contain a list of chapter beginning timestamps $B$ and chapter titles $D$. We formulate the video chapter generation into a two-stage framework. Figure 2 demonstrates the two-stage framework: In the first stage, A two-stream network with a sliding window mechanism is developed for chapter localization. Then, reused visual model and fusion module are applied for multi-modal feature extraction, feature fusion and title generation.

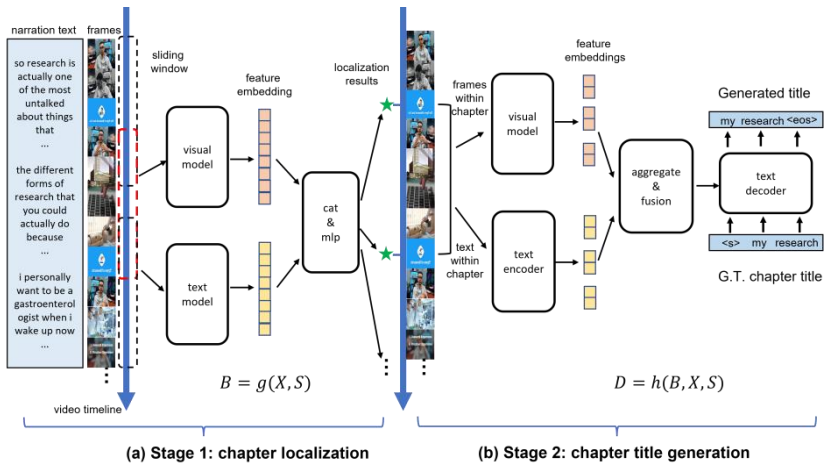

Figure 2: The framework of our approach. (a) Chapter localization stage. A two-stream network with a sliding window mechanism. (b) Chapter title generation stage. The Transformer encoder-decoder with reused visual model and feature fusion module is used for generating chapter title.

## 4.1 Chapter Localization

This stage takes into account the video frames $X$ and narration text $S$ to localize every chapter beginning time in a specific second. To parse the whole of video, the sliding window mechanism can be applied to determine where the chapter beginning time is. Note that our sliding window mechanism is different from the ones in temporal action localization [17, 26]. Instead of using sliding windows to fit target intervals [26], we use sliding windows to gather information and decide if the chapter begins within a time window.

The video context in a time window can be defined as a video clip. Each clip contains some images $X_{t_a} = \{x_t\}_{t=a}^{a+k}$ and narration words $S_{t_a} = \{s_t\}_{t=a}^{a+k}$, where $k$ is the size of time window and $a$ is the clip start time. Given a clip context $X_{t_a}$, $S_{t_a}$, we would like to evaluate the chapter beginning probability $p_{t_a}$. We formulate chapter localization function as a binary classifier $p_{t_a} = g(X_{t_a}, S_{t_a})$. The binary logistic regression loss function $L_b$ can be defined for training:

$$L_b = \sum_{t_a} (l_{t_a} \cdot \log(p_{t_a}) + (1 - l_{t_a}) \cdot \log(1 - p_{t_a})) \quad (1)$$

where $t_a$ denotes the clip start time. $l_{t_a}$ is binary 0-1 label which represents if a chapter begins in a clip. Negative data are dominated in all clips. We remedy this imbalance distribution by oversampling positive clips during training. Moreover, we need to localize the chapter beginning time in a specific second. To fulfill it, we simply use the middle time of positive clips as the final localization results $\{b_i\} \in B$:

$$b_i = t_a + \frac{k}{2} \quad (2)$$

The Eq. 1 provides a training target to decide where the chapter beginning time is. Nevertheless, sliding clip windows second by second is inefficient, especially when processing long-duration videos. Since the video in ChapterGen dataset consists of hundreds and thousands of seconds of content. A few seconds offset in localization result is negligible in practice. We allow a small time offset $o$ round the groundtruth chapter beginning time $\hat{t}$. All $\hat{t}$ can be sampled by sliding windows when setting the sliding step $\leq 2o$. We adopt the maximum $u = 2o$ as the skip sliding step to achieve the best efficiency without losing any potential chapter beginning times. For more details about skip sliding mechanism, please refer to supplementary material.

## 4.2 Chapter Title Generation

The localization results from the first stage provide chapter boundaries that can be used for selecting relevant video content. Based on those relevant video content, the model generates chapter titles. To achieve the generation ability, we adopt the standard Transformer encoder-decoder architecture [33]. the goal of encoder-decoder Transformer is to estimate the conditional probability $p(d_1, ..., d_{n'}|s_1, ..., s_n)$ where $(s_1, ..., s_n)$ is the narration text sequence within a chapter boundary and $(d_1, ..., d_{n'})$ is its corresponding chapter title sequence whose length $n'$ could much shorter than $n$. For fusing the text and language features without modifying encoder-decoder architecture, one possible way is to apply concatenation with MLP structure. However, simple concatenation may not be able to capture long-range interaction between text and visual features. So we design a cross attention fusion layer to integrate these two modalities.

Specifically, the trained visual model from the first stage is reused and outputs a series of visual embeddings $r_{1:m}$ which describes the visual dynamics within a chapter. The Transformer encoder extracts text embeddings $u_{1:n}$ from a chapter of video narration. Suppose the size of a visual embedding is $p$ and text embedding is $q$. The cross attention fusion layer first maps all visual embeddings (i.e. a embedding matrix) $R_{m \times p}$ to a key matrix $K_{m \times q}$ and a value matrix $V_{m \times q}$. All text embeddings $U_{n \times q}$ are mapped as a query matrix $Q_{n \times q}$. Then, fused output $U'_{n \times q}$ is calculated by following the standard attention mechanism:

$$U'_{n \times q} = softmax(Q_{n \times q} * K^T_{m \times q}) * V_{m \times q} + U_{n \times q} \qquad (3)$$

Note that the attended results are added by original text embeddings $U_{n \times q}$ as a skip connection to remedy gradient vanish.

Since the dimension of fused embeddings $U'_{n \times q}$ is the same with text encoder results $U_{n \times q}$, we can directly leverage the standard decoder to generate target words one by one:

$$p(d_1, ..., d_{n'}) = \prod_{i=1}^{n'} p(d_i | U'_{n \times q}, d_1, ..., d_{i-1}) \qquad (4)$$

where every $p(d_i|U'_{n \times q}, d_1, ..., d_{i-1})$ distribution can be calculated by using a softmax over all the words. We initialize the encoder and decoder with pretrained parameters on large text corpora [36, 38] and finetune the model on ChapterGen dataset to generate chapter titles.

## 5 Experiments

In this section, we conduct several experiments: (1) Evaluation of models' localization ability, (2) Evaluation of models' summarization ability, (3) Case study, (4) Ablation study (supplementary materials).

## 5.1 Experiment Setup

**Dataset Details.** The ChapterGen dataset is uniformly split at random to train, validation and test subsets with 6742, 963, 1926 samples respectively. The fraction of easy/hard on training, validation and testing sets is about 2.7/1 as shown in Table 3.

| Dataset | Videos | Clips | Frames | Words |
|---|---|---|---|---|
| Train | 6,742 | 1,346,696 | 5,484,523 | 15,794,075 |
| Validation | 963 | 196,087 | 798,287 | 2,302,973 |
| Test | 1,926 | 388,743 | 1,583,021 | 4,551,809 |

Table 3: The statistics of train, validation and test split.

**Evaluation Metrics.** For chapter localization, we follow the metrics **AP** (Average Precision) and **Recall@c** [5, 22] for the sake of fair comparison. The Recall@c calculates the percentage of the ground truth chapter start time that is within $c$ seconds of the predicted ones. For chapter title generation, we adopt the commonly used metrics **ROUGE F1** scores (i.e. ROUGE-1, ROUGE-2, ROUGE-L) in text summarization [21, 39].

**Implementation Details.** The visual feature is processed by standard pretrained ResNet-50 [8] with TSM component [16]. Compared with 3D CNN [32], the TSM can be seamlessly integrated with 2D CNN to capture temporal features without introducing heavy computation burdens. The narration text feature is extracted by BERT [6]. Before training, we finetune the BERT on ChapterGen dataset by unsupervised Masked Language Model (MLM) task.

In chapter title generation, we apply Transformer encoder-decoder [33] and initialize it with pretrained Pegasus [38] and Bigbird [36]. The maximum generated title length is set to 30 words. In cross attention fusion module, the length of feature sequence is set to $n = 512, m = 10$ for language and visual modalities. The feature sequence can roughly cover hundreds of seconds video content. We train all models end-to-end using AdamW optimizer [19] with batch size 64. The learning rate is set to $1e^{-5}$ with linear warmup and cosine decay scheduler.

## 5.2 Main Results

**Chapter Localization.** As shown in Table 4, our method with both visual and text modalities achieves the best results in AP and Recall. This verifies the effectiveness of our model. Our method with only text modality achieves the fastest inference speed and better performance compared with the visual model. It indicates that the narration text contains more discriminative features than video frames in chapter localization. The visual clues are complementary for text and combination of visual and text features can further boost results (AP 38.5% -43.3%).

[5, 22] are the state-of-the-art methods on movie scene segmentation and BMT [11] is a recent dense video captioning approach which takes visual I3D features and audio VGG features to generate event proposal and text description. As shown in Table 4, our method with visual modality outperforms them, which indicates that our strategy is more straightforward and effective on chapter localization task.

**Chapter Title Generation.** For chapter title generation, we compare methods among BMT[11], two SOTA text summarization approaches: Pegasus [38], Bigbird [36] and three heuristic baselines which are widely applied in text summarization [28, 38, 39]. **Random:**

| Method | Modality | Inference FPS | AP | Recall (0.5 thr.) | Recall@3s (0.5 thr.) | Recall@5s (0.5 thr.) | Method | Modality | Inference FPS | AP | Recall (0.5 thr.) | Recall@3s (0.5 thr.) | Recall@5s (0.5 thr.) |
|---|---|---|---|---|---|---|---|---|---|---|---|---|---|
| Random | - | - | - | 1.1 | 7.4 | 11.1 | Ours | visual | 2874.2 | 26.7 | 23.2 | 52.0 | 67.0 |
| [22] | visual | 2652.8 | 6.6 | 8.7 | 18.9 | 21.1 | Ours | text | 5934.0 | 38.5 | 23.4 | 56.2 | 71.3 |
| [5] | visual (fz.) | 2740.0 | 7.9 | 13.1 | 33.2 | 38.6 | Ours | visual + text | 1156.9 | **43.3** | **25.8** | **60.1** | **76.1** |
| [5] | visual (unfz.) | 2740.0 | 8.7 | 15.9 | 38.3 | 43.7 | - | - | - | - | - | - | - |
| [11] | visual + audio | - | 23.4 | 19.8 | 52.8 | 59.9 | - | - | - | - | - | - | - |

Table 4: The chapter localization results. The Inference FPS is the average model inferred frames per second which indicates the model efficiency. The bold fonts show the best results.

| Method | Modality | Setting | ROUGE-1 | ROUGE-2 | ROUGE-L |
|---|---|---|---|---|---|
| Random | text | All | 4.9/5.0 | 0.9/1.0 | 4.7/4.7 |
| Lead | text | All | 10.7/10.0 | 4.1/3.6 | 10.3/9.6 |
| Principal | text | All | 9.4/8.9 | 3.4/3.0 | 9.0/8.4 |
| BMT [11] | visual+audio | All | 18.7/13.7 | 7.2/3.1 | 18.6/13.8 |
| Ours (B.) | text | All | 16.6/11.6 | 4.9/3.1 | 16.3/11.3 |
| Ours (P.) | text | All | 26.9/19.7 | 8.4/5.6 | 26.6/19.4 |
| Ours (P.) | visual+text cat | All | 28.0/20.3 | 8.4/5.5 | 27.8/20.0 |
| Ours (P.) | visual+text cross | All | **34.4/25.6** | **13.4/8.8** | **34.0/25.3** |
| BMT [11] | visual+audio | Easy | 19.9/14.8 | 8.0/4.0 | 19.8/15.0 |
| Ours (B.) | text | Easy | 18.9/12.6 | 6.2/3.7 | 18.5/12.3 |
| Ours (P.) | text | Easy | 29.7/20.9 | 10.4/6.5 | 29.4/20.7 |
| Ours (P.) | visual+text cat | Easy | 31.5/22.1 | 10.7/6.4 | 31.2/21.8 |
| Ours (P.) | visual+text cross | Easy | **38.9/28.3** | **17.0/10.4** | **38.6/28.0** |
| BMT [11] | visual+audio | Hard | 14.5/11.0 | 2.7/2.2 | 14.8/11.8 |
| Ours (B.) | text | Hard | 13.2/9.8 | 3.2/2.3 | 12.9/9.6 |
| Ours (P.) | text | Hard | 22.2/17.5 | 5.2/3.9 | 21.9/17.2 |
| Ours (P.) | visual+text cat | Hard | 22.9/17.9 | 4.8/4.0 | 22.6/17.6 |
| Ours (P.) | visual+text cross | Hard | **27.3/21.3** | **7.6/6.0** | **27.0/21.0** |

Table 5: The chapter title generation results (G.T. chapter locations/Predicted chapter locations). Our(B.) and Our(P.) represent the Bigbird and Pegasus pretrain respectively. In modality, cat and cross denote concatenation fusion and cross attention fusion respectively. All, easy and hard settings represent all, easy case and hard case testing data.

uniformly select one sentence (i.e. 10 consecutive words) at random from the original narration text. **Lead:** select the first 10 words from the narration text as the chapter title. **Principal:** select top-1 scored sentence based on importance. The importance score is calculated by ROUGE-1 F-Score between the sentence and the rest of the narration text.

As shown in Table 5, Lead strategy is the best among three heuristic strategies. This may suggest that the first couple of sentences in each chapter contain the most conclusive information. BMT [11] outperforms all heuristic strategies but it is worse than Ours(P.) with text only. Compared with narration text, the audio information may contain noises like BGM, ambient sound especially the long video in chapter generation. As a result, our method with only text modality outperforms BMT.

The model Our(P.) (B.) without visual capturing module can be treated as a standard text summarization process. When introducing visual modality, the performance can be further improved (i.e. Our(P.) with visual and text modalities). To show the effectiveness of cross attention fusion, we implement feature concatenation with MLP as a fusion baseline. The cross attention (i.e. visual+text cross) surpasses concatenation (i.e. visual+text cat) with a large margin.

Moreover, to better illustrate the performance of our models' summarization ability, we feed the summarization module with G.T. chapter locations (left) which can eliminate the negative effect of chapter localization module. This suggests that the accuracy of chapter localization could affect the performance of title chapter generation in a large degree and prove the importance of accurate chapter localization for high quality title generation.

## 5.3 Ablation Study

We present ablation studies for the chapter localization stage to verify the choices of different sliding window size hyperparameters $k$, the influence of the TSM component, and the results on easy and hard cases.

**Choice of sliding window size $k$.** The sliding window size determines how much context information is considered for chapter localization. The small window size is helpful for the model to focus on most related content. The large window size provides more plentiful content information but may introduce more noises and extra computation burden. Figure 3 shows the results of our two-stream model with different sliding window sizes. Window size $k = 16$ achieves best trade-off.

**Influence of TSM**. We capture visual features in temporal by using TSM [16]. Compared with 3D CNN [32], TSM is lightweight and can be seamlessly integrated with 2D CNN. The performance comparison is shown in Figure 3. The visual model with TSM outperforms no TSM (AP 26.7% vs. 21.1%).

**Results on easy and hard cases**. As illustrated in Figure 1, some videos have clear content and temporal structures, while some are vague. We conduct experiments on easy, hard and all cases to investigate how different input data influence the performance. As shown in Figure 3, the results on easy cases are better than hard cases. When mixing up all easy and hard cases, the performance is in-between easy and hard. The similar conclusion can also be observed for chapter title generation in Table 5.

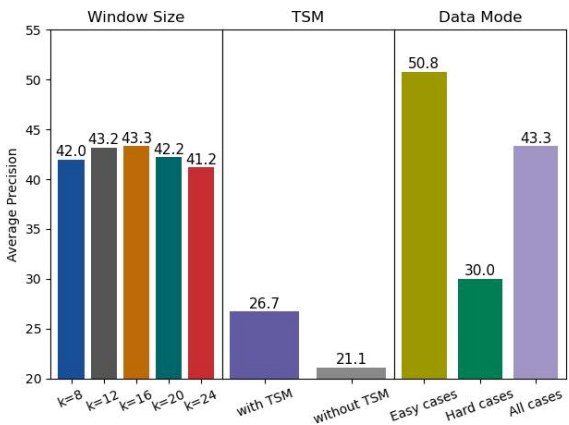

Figure 3: The ablation study results on different settings.

411## 5.4 Case Study

We present qualitative results on both chapter localization and chapter title generation to further explore their effectiveness. Figure 4(a) visualizes the comparison between G.T. chapters and generated chapters. Our framework can localize the chapter beginning time and generate desirable chapter titles.

Figure 4(b) demonstrates the saliency map on both video frames and narration text. It is a clip of images and narration text in a sliding window. We can observe that the regions of sugar, stomach and tobacco contribute the most in the visual modality. And "number 18", "tobacco", "mechanism" keywords in narration text are captured as important features. Based on this saliency map, we come to some empirical conclusions: (1) The visual model captures different regions (i.e. background or foreground) on different frames. It can focus on relevant regions to extract visual features. (2) The text model can grip attention on some important keywords for prediction.

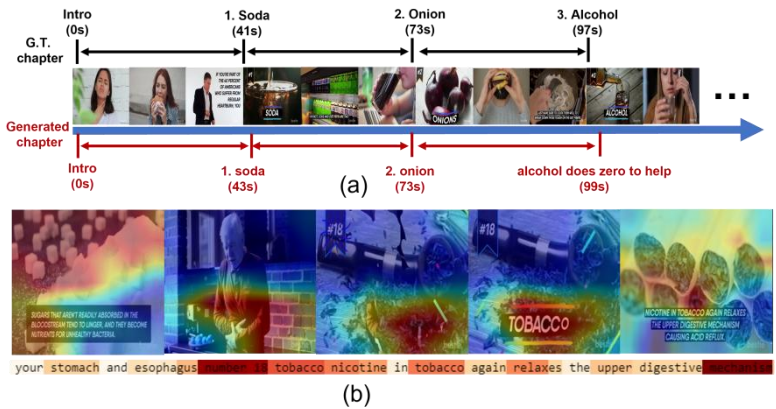

Figure 4: The chapter generation results and saliency maps. Our framework can precisely localize the chapter start time and generate semantically reasonable chapter titles.

## 6 Conclusion and Future Work

In this work, we propose a new high-level video understanding task: video chapter generation. For facilitating the research in this task, we construct a new dataset ChapterGen, which is collected from readily available resources without expansive manual annotation. To address this task, a two-stage framework is developed. The proposed framework contains chapter localization and chapter title generation. In the future, we would like to expand our dataset to larger scale so as to improve the generalization.